\pdfoutput=1

\documentclass[11pt]{article}

\usepackage[]{acl}

\usepackage{times}
\usepackage{latexsym}

\usepackage[T1]{fontenc}

\usepackage[utf8]{inputenc}

\usepackage{microtype}
\usepackage{enumitem}
\usepackage{graphicx}
\usepackage[export]{adjustbox}
\usepackage{amsmath}

\usepackage{array}
\usepackage{enumitem}
\usepackage{amsmath,amssymb}
\usepackage{colortbl}
\usepackage{multirow}
\usepackage{booktabs}
\usepackage{pifont}
\usepackage{pgfplots}
\usepackage{collcell}
\usepackage{siunitx}
\newcommand{\benchmark}{\textsc{GlobalRG}}

\title{From Local Concepts to Universals:\\ Evaluating the Multicultural Understanding of Vision-Language Models}

\author{Mehar Bhatia ~~Sahithya Ravi ~~Aditya Chinchure ~~Eunjeong Hwang ~~Vered Shwartz\\
University of British Columbia\\Vector Institute for AI\\
{\tt \{meharb23, vshwartz\}@cs.ubc.ca} \\ \\
\href{https://globalrg.github.io/}{\tt globalrg.github.io/}}

\begin{document}
\maketitle
  
\begin{abstract}
Despite recent advancements in vision-language models, their performance remains suboptimal on images from non-western cultures due to underrepresentation in training datasets. Various benchmarks have been proposed to test models' cultural inclusivity, but they have limited coverage of cultures and do not adequately assess cultural diversity across universal as well as culture-specific local concepts. To address these limitations, we introduce the \benchmark{} benchmark, comprising two challenging tasks: \textit{retrieval across universals} and \textit{cultural visual grounding}. The former task entails retrieving culturally diverse images for universal concepts from 50 countries, while the latter aims at grounding culture-specific concepts within images from 15 countries. Our evaluation across a wide range of models reveals that the performance varies significantly across cultures -- underscoring the necessity for enhancing multicultural understanding in vision-language models.

\end{abstract}

\section{Introduction}
\label{sec:intro}
Vision-Language Models (VLMs) have shown emergent capabilities through large-scale training that have made them gain popularity in recent years. VLMs show promising results across various vision and language tasks, from image captioning to visual question answering and cross-modal retrieval and grounding. A key component contributing to their strong performance across the board is the scale of their pre-training datasets. However, these large-scale datasets tend to predominantly contain images from Western cultures \cite{shankar2017no}. The underrepresentation of certain cultures in the data translates into performance disparities across cultures.  \cite{de2019does,gustafson2023pinpointing}.

Several benchmarks and datasets have been proposed to test the cultural inclusivity of VLMs. These include testing the models' performance on questions pertaining to images from certain cultures \cite{liu-etal-2021-visually,yin-etal-2021-broaden}, on their ability to adapt images from one culture to another \cite{khanuja2024image}, or on stereotypical depiction of various cultures \cite{jha2024beyond}. Nonetheless, existing benchmarks address a limited set of cultures (5-7), leaving a substantial representational gap. Moreover, current benchmarks leave out a crucial aspect: assessing the cultural diversity in the representation of universal concepts. 

\begin{figure}[t]
    \centering
    \includegraphics[width=.45\textwidth]{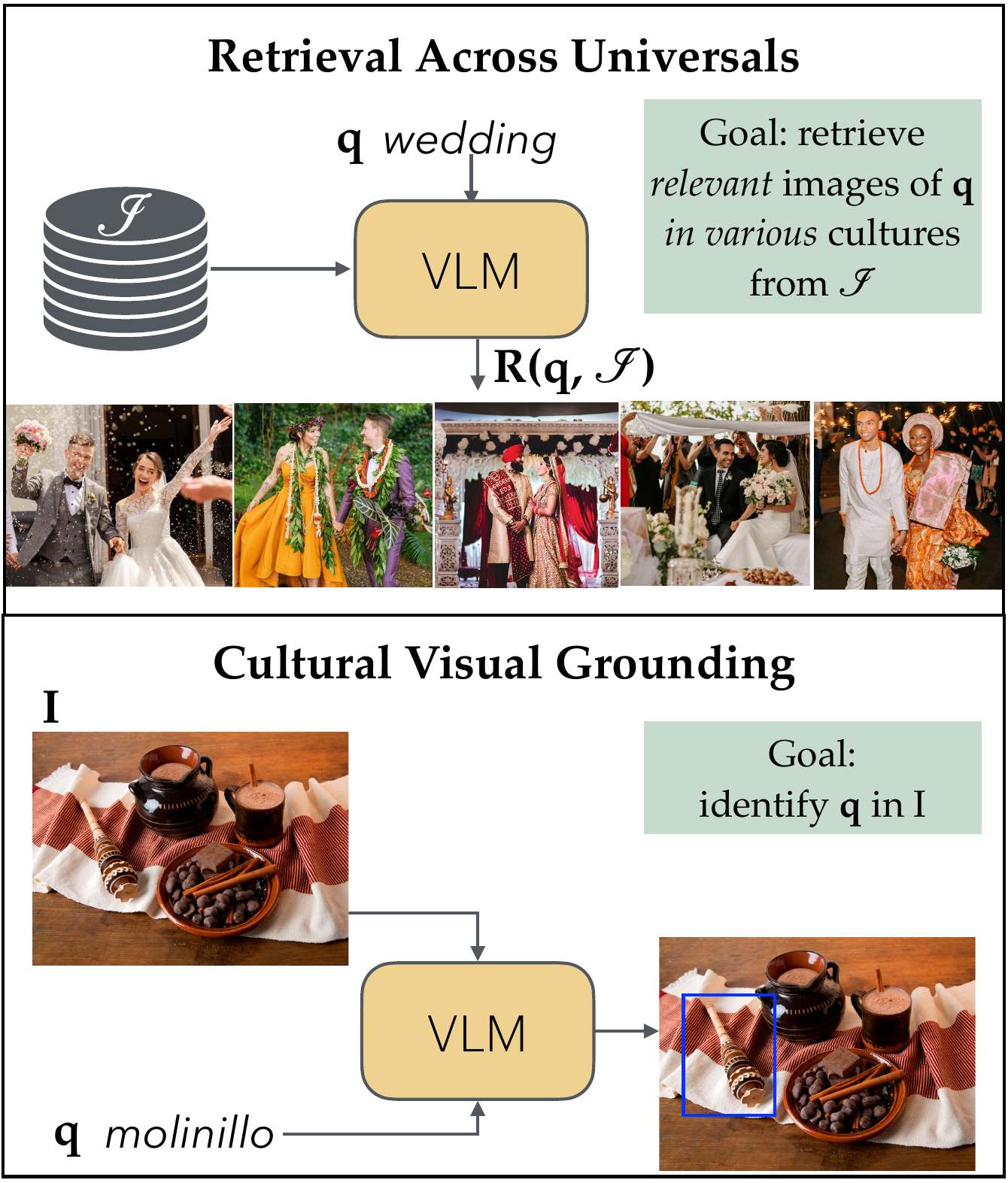}
    \caption{An example instance from each task in \benchmark{}: i) \emph{Retrieval Across Universals} measures the ability of VLMs to retrieve culturally diverse images for a query q. ii) \emph{Cultural Visual Grounding} aims to evaluate the ability of VLMs to identify a cultural concept q. }
    \label{fig:motivation}
\end{figure}

To address this gap, we present the \benchmark{} benchmark, which consists of two tasks (Figure~\ref{fig:motivation}). The first task, \textbf{retrieval across universals}, covers images from 50 countries across 10 regions. It assesses the ability of VLMs to retrieve culturally-diverse images pertaining to textual prompts of universal concepts such as ``breakfast'' and ``wedding’’. In addition to the standard precision@k metric, which verifies that the retrieved images correctly depict the target concept, we also propose a new metric, diversity@k, that measures the cultural-diversity among the retrieved images, allowing us to identify models' bias towards specific countries or regions. 

In the second task, \textbf{cultural visual grounding}, we cover 15 countries across 8 regions and evaluate models' ability to ground culture-specific concepts (e.g., ``molinillo'', Mexican whisk) within an image. 

Extensive evaluation on 7 models for the retrieval task and 5 models for the grounding task reveals discrepancies across cultures, reassessing findings by prior work \cite[e.g.,][]{liu-etal-2021-visually,yin-etal-2021-broaden}. We further analyze whether VLMs exhibit biases towards certain cultures. In the grounding task, the performance on North America and Europe is substantially higher than on East Asia and South East Asia. This preference is inconsistent across universals in the retrieval task, e.g., a model may retrieve European images of funerals but African images of farming. A closer look reveals that even when models retrieve seemingly diverse images, they often share Western elements, such as eggs for breakfast and white dresses at weddings.  

\benchmark{} highlights the lack of cultural awareness in current VLMs. By identifying and addressing these gaps, we can work towards developing models that perform equally well on inputs pertaining to concepts and images from diverse cultures.

\section{Related Work}
\label{sec:related_work}
\paragraph{The Geo-Diversity Problem.} Existing large-scale vision and language datasets are imbalanced in their representation of different regions, over-representing the West \cite{shankar2017no}. As a result, models trained on these datasets may exhibit discrepancies in performance when introduced with inputs concerning various demographic and geographic factors \cite[e.g.][]{gustafson2023pinpointing, de2019does}. For instance, image generation models---when asked to generate images of universal concepts such as ``house'', tend to depict the concept as it appears in the US or India, cultures that are more prominently featured in the training data \cite{basu2023inspecting}.

To serve users from diverse cultures fairly, it is imperative to collect large-scale datasets from diverse data sources
\cite{kim2021towards, goyal2022vision}. Two recent geo-diverse image datasets that are popular for training geo-diverse VLMs, Dollar Street \cite{rojas2022dollar} and GeoDE \cite{ramaswamy2024geode}, focus on common household items, lacking coverage of more abstract and culture-specific concepts. Finally, to make cross-cultural data collection more feasible, researchers proposed to apply domain adaptation \cite{kalluri2023geonet} and active learning \cite{ignat2024annotations} based on visual similarity. 

\paragraph{Geo-Diverse Benchmarks.} With the understanding that language has a social function, there has been growing interest in the NLP community in making models more culturally inclusive \cite[e.g.,][]{hershcovich-etal-2022-challenges,nguyen2023extracting,bhatia-shwartz-2023-gd}. Several benchmarks have been developed to test language models' cultural awareness with respect to values and social norms \cite{durmus2023towards}, culinary norms \cite{palta-rudinger-2023-fork}, figurative language \cite{kabra-etal-2023-multi}, and more. 

In the multimodal domain, benchmarks have been developed to test VLMs on visual question answering and reasoning \cite{liu-etal-2021-visually,yin-etal-2021-broaden,zhou2022vlue}, image-text retrieval and visual grounding \cite{zhou2022vlue}, image captioning \cite{ye2023cultural}, and cultural adaptation \cite{khanuja2024image}. 

Despite these efforts, current benchmarks typically cover an incredibly small number of cultures (5-7). To bridge this gap, we introduce a benchmark with two tasks covering 50 and 15 cultures respectively. Moreover, our benchmark tests models both on their familiarity with \emph{culture-specific} concepts and on the diversity of their representation of \emph{universal concepts}.

\section{Task 1: Retrieval across Universals}
\label{sec:task1}
\begin{table}[t]
\centering
\scriptsize
\setlength{\tabcolsep}{2.2pt}
\begin{tabular}{lp{5.3cm}}
\toprule
\textbf{Region} & \textbf{Countries} \\
\midrule
\textbf{East Asia}               & China, South Korea, Japan \\
\textbf{South East Asia}         & Vietnam, Thailand, Philippines, Indonesia, Singapore \\
\textbf{South Asia}              & India, Pakistan, Sri Lanka \\
\textbf{Middle East Asia}        & Saudi Arabia, Iran, Turkey, Lebanon, Egypt \\
\textbf{Europe }                 & Italy, Greece, France, Germany, Netherlands, Portugal, Spain, United Kingdom, Poland, Sweden, Hungary, Bulgaria, Russia \\
\textbf{Africa }                 & Tanzania, Kenya, Uganda, Ghana, Nigeria, Ethiopia, South Africa, Morocco, Tunisia \\
\textbf{Latin America}           & Brazil, Peru, Chile, Argentina, Mexico \\
\textbf{Caribbean }              & Jamaica \\
\textbf{Oceania}                 & Australia, New Zealand, Fiji \\
\textbf{North America}           & USA, Canada \\
\bottomrule
\end{tabular}
\caption{List of cultures covered in the retrieval task.}
\label{tab:retrieval-cultures}
\end{table}

Image-text retrieval is a fundamental task for evaluating VLMs, where the objective is to retrieve relevant images based on textual queries. 
Existing retrieval benchmarks such as COCO \cite{lin2014microsoft}, Flicker30K \cite{plummer2015flickr30k}, ImageCoDe \cite{krojer-etal-2022-image}, and CIRR \cite{liu2021image} contain images predominantly from North America and Europe. To develop globally effective retrieval systems, it is crucial to evaluate models on culturally heterogeneous datasets. In this work, we present a dataset containing images from 50 cultures (Table~\ref{tab:retrieval-cultures}). We introduce the novel task of \textbf{Retrieval across Universals}, aimed at retrieving culturally-diverse images for universal concepts such as ``wedding''. We describe the dataset collection in Sec~\ref{sec:retrieval:dataset}. 

Image-text retrieval is typically evaluated using precision.
Beyond measuring the correctness of the retrieved images, this metric overlooks a significant aspect of retrieval systems: \emph{cultural diversity}. We thus propose an additional evaluation metric to measure the cultural diversity of the retrieved images (Sec~\ref{sec:retrieval:eval_metric}). We evaluate an extensive number of VLMs on the retrieval task (Sec~\ref{sec:retrieval:baselines}) and report the results in Sec~\ref{sec:retrieval:results}. 

\begin{table}[t]
\centering
\scriptsize
\tt
\begin{tabular}{lll}
\toprule
breakfast & clothing & dance \\
dessert & dinner & drinks \\
eating habits & farming & festival \\
funeral & greetings & head coverings \\
instrument & lunch & marriage \\
music & religion & ritual \\
sports & transport \\
\bottomrule
\end{tabular}
\caption{Human universals used as textual queries in our retrieval dataset.}
\label{tab:universals-list}
\end{table}

\subsection{Dataset Collection}
\label{sec:retrieval:dataset}

\paragraph{Textual Queries.} The queries in our dataset are human universals---concepts common across cultures worldwide, such as ``clothing'' and ``dance''. Table~\ref{tab:universals-list} presents the list of 20 human universals used as textual queries in our dataset. The list was adapted from an extensive list of 369 human universals by \newcite{brown2004human} and \newcite{pinker2004blank}. We manually selected human universals that can be depicted in images. For example, universals like ``clothing'' are associated with tangible objects, and ``dance'' is a ritual that can be visually depicted. In both cases, these universal concepts are expected to be visually represented differently across diverse cultures.\footnote{The complete list of human universals can be found here: \href{https://condor.depaul.edu/~mfiddler/hyphen/humunivers.htm}{https://condor.depaul.edu/$\sim$mfiddler/hyphen/humunivers.htm}} 

\paragraph{Images.} To obtain culturally diverse images corresponding to the textual queries, we first used CANDLE \cite{nguyen2023extracting}, a comprehensive corpus of cultural knowledge, to extract 3 sentences corresponding to each universal concept and each culture. For example, for ``wedding'' and ``India'', CANDLE contains the sentence ``\textit{The mehendi ceremony holds significance in Indian tradition}''. These sentences provide context and cultural specificity for each universal. We use these sentences to scrape images from Google Images. To ensure the quality of the images, one of the authors manually verified each image in the dataset, filtering out low-resolution images, images with text, and images depicting multiple scenes (i.e., grid images). The final dataset includes a total of 3,000 visually-diverse images (50 cultures $\times$ 20 universals $\times$ 3 images).

\subsection{Task Definition and Evaluation Setup}
\label{sec:retrieval:eval_metric}

\begin{table*}[t]
\centering
\scriptsize
\setlength{\tabcolsep}{2.2pt}
\begin{tabular}{llrcccccc}
\toprule
 \textbf{Model} & \textbf{Training Data} & \textbf{Data Size} & \multicolumn{2}{c}{\textbf{Relevance}}  &  \multicolumn{2}{c}{\textbf{Diversity (Country)}} & \multicolumn{2}{c}{\textbf{Diversity (Region)}}\\
 
\cmidrule(lr){4-5}
\cmidrule(lr){6-7}
\cmidrule(lr){8-9}
 &  &  & \textbf{prec@5} & \textbf{prec@10} & \textbf{div@5} & \textbf{div@10} & \textbf{div@5} & \textbf{div@10} \\
 
\midrule
\multicolumn{3}{l}{\underline{\textbf{Dual-Encoder:}}} \\ 
CLIP \cite{radford2021learning} & web-scraped & 400M & 72.5 & 70.0 & 93.96 & 94.16 & 66.71 & 64.64  \\
OpenCLIP \cite{cherti2023reproducible} & LAION-2B & 2B & 69.5 & 75.0 & 95.69 & 95.14 & 73.39 & \textbf{66.93}\\

\midrule
\multicolumn{3}{l}{\underline{\textbf{Encoder-Decoder:}}} \\ 
CoCA \cite{yu2022coca} & JFT-3B & 3B & \textbf{81.0} & \textbf{79.5} & \textbf{98.27} & \textbf{95.37} & 68.18 & 64.88\\

\midrule
\multicolumn{3}{l}{\underline{\textbf{Dual Encoder + Multimodal Fusion Encoder:}}} \\ 
TCL \cite{yang2022vision} & CC-3M, SBU, COCO, VG & 4M & 76.0 & 74.5 & 92.78 & 91.22 & 74.04 & 66.54\\
ALBEF \cite{li2021align} & CC-12M, SBU, COCO, VG & 14M & 68.0 & 70.0 & 92.24 & 91.11 & 65.75 & 64.63 \\
BLIP2 \cite{li2023blip} & CC-3/12M, SBU, COCO, VG, LAION-115M & 129M & 74.0 & 74.5 & \textbf{98.27} & 92.96 & \textbf{74.25} & 63.26 \\
FLAVA \cite{singh2022flava}  & CC-3/12M, SBU, COCO, VG WIT, Red Caps, YFCC & 70M & 60.0 & 62.0 & 96.54 & 94.95 & 72.32 & 66.84\\ 
\bottomrule
\end{tabular}
\caption{Average performance of various VLMs on the the retrieval across universals task, in terms of \textbf{Relevance} and \textbf{Diversity}.}
\label{tab:models-retrieval}
\end{table*}

We introduce the novel task of \textbf{Retrieval across Universals}, aimed at retrieving culturally diverse images for a given universal concept. Formally, let \( \mathcal{Q} = \{q_1, q_2, \ldots, q_n\} \) be a set of textual queries representing universal concepts, and \( \mathcal{I} = \{I_1, I_2, \ldots, I_m\} \) the set of images from different cultures. Given a query \( q \in \mathcal{Q} \), the goal is to retrieve a ranked list of images \( \mathcal{R}(q, \mathcal{I}) = \{I_{r_1}, I_{r_2}, \ldots, I_{r_k}\} \subset \mathcal{I} \) that maximizes both relevance and cultural diversity.

\begin{itemize}[nosep,leftmargin=10pt,labelindent=*]
    \item \textbf{Relevance}: \( \text{Rel}(q, I) \) refers to how well the image \( I \) matches the query \( q \). 
    \item \textbf{Diversity}: \( \text{Div}(\mathcal{R}(q, \mathcal{I})) \) measures the cultural diversity of the retrieved images. 
\end{itemize}

Specifically, relevance is captured by the standard precision@k, the ratio of the top k retrieved images that correctly answer the query. For diversity, we propose the diversity@k metric, which uses entropy to measure the cultural diversity among the top k retrieved images: \begin{equation}
    \textit{diversity }@k = -\frac{1}{\log\left(\frac{1}{m}\right)} \sum_{i=1}^{m} p_i \log(p_i)
\label{eq:diversity}
\end{equation}

\noindent where \( p_i \) is the proportion of images from the \( i \)-th culture in the top k retrieved images $\mathcal{R}(q)$, and \( m \) is the total number of cultures in the top k. A high normalized entropy value ($\sim 1$) indicates high diversity, meaning the retrieved images are well-distributed across different cultures. Conversely, a low entropy value ($\sim 0$) indicates low diversity, suggesting that the retrieved images are biased towards specific cultures.  We report diversity with respect to both the country and the region.

Our balanced focus on relevance and diversity ensures that models are evaluated not only on their ability to understand and represent concepts accurately but also on their capacity to do so across cultures. 

\subsection{Models}
\label{sec:retrieval:baselines}

We evaluate the performance of several state-of-the-art VLMs on the retrieval task. The models are categorized based on their architectural design and training methodologies in Table~\ref{tab:models-retrieval}. We cover a diverse set of models, including dual encoder and encoder-decoder, as well as dual encoders with multimodal fusion encoder. These models facilitate cross-modal alignment via a multitude of pre-training objectives, including contrastive loss on uni-modal encoders, image-text matching, masked language modelling, and more.\footnote{We could not evaluate advanced closed-source models like GPT-4v or Gemini on our retrieval task since these models do not support searching through our large collection of images.}

\begin{figure*}[t]
     \centering
     \includegraphics[width=\textwidth]{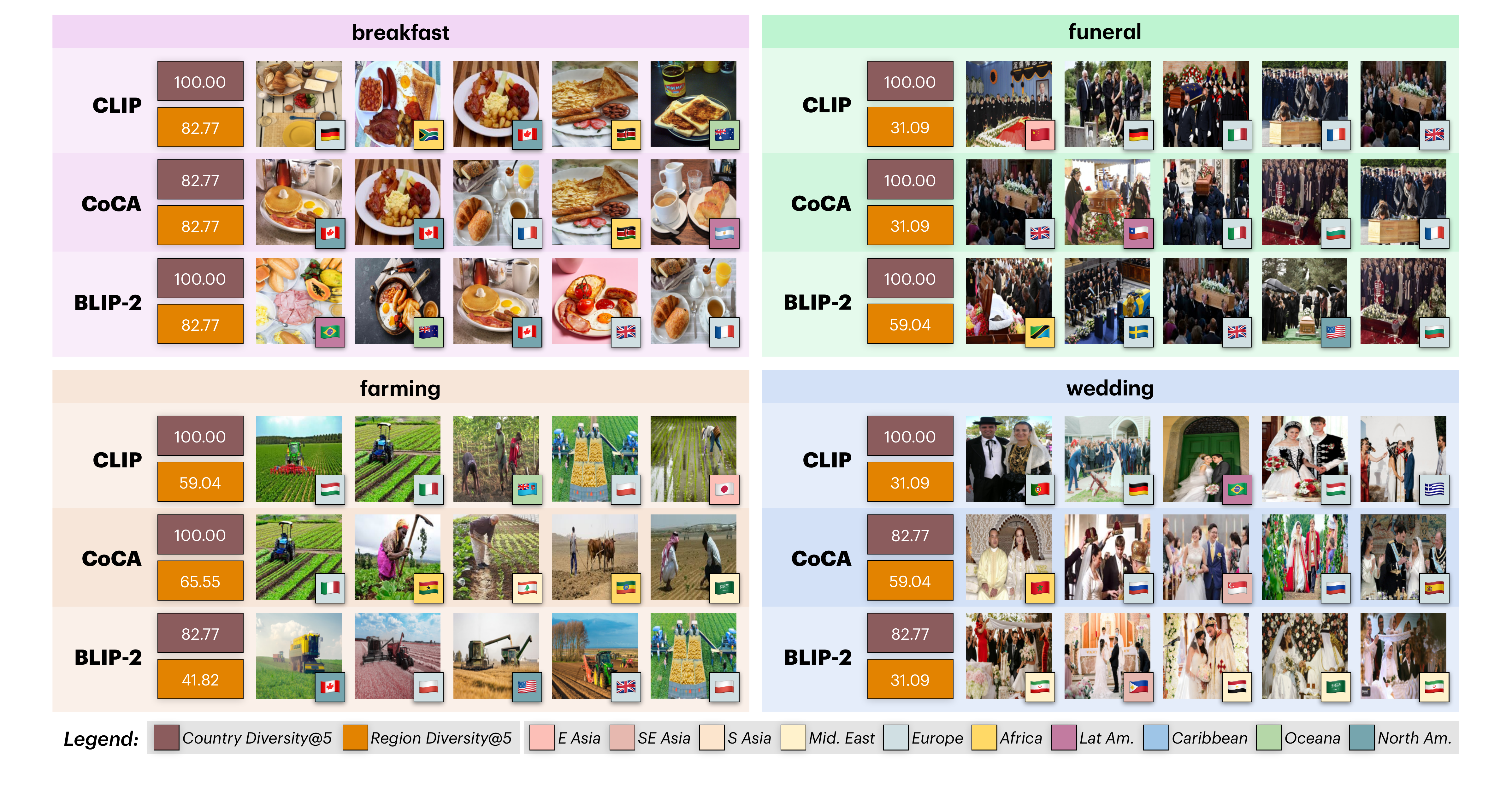}
     \caption{Top 5 images retrieved for a sample of the universals by models CLIP, CoCA and BLIP-2. Each image is annotated with a flag representing the country, and the background colour of the flag represents the region.} 
     \label{fig:retrieval-analysis}
 \end{figure*}

\subsection{Results and Analysis}
\label{sec:retrieval:results}

\paragraph{RQ$_1$: Are VLMs able to retrieve relevant and culturally diverse images for universal concept words?} Table~\ref{tab:models-retrieval} presents the relevance and diversity scores for each model (see Appendix~\ref{appendix:results_metrics} for a complete breakdown by universal). With respect to relevance, models achieve moderate to high precision scores, with CoCA leading by 5 points. 

We note that country-level diversity scores are high for all models, indicating that VLMs can retrieve images from a variety of geographical contexts. Among them, CoCA performs exceptionally well, likely attributed to its extensive training on 3 billion images from Google's proprietary JFT dataset \cite{zhai2022scaling}. 

Similarly, in dual-encoder models, OpenCLIP demonstrates superior cultural diversity, benefiting from its large training dataset of 2 billion images.  CLIP, which uses the same dual-encoder architecture and contrastive loss objectives as OpenCLIP but is trained on a dataset five times smaller, exhibits lower performance across all metrics. Naturally, pre-training on a larger-scale dataset increases the chances that the model was exposed to more culturally diverse images. In contrast, regional diversity scores are notably lower across the board. At the same time, for country diversity@5, BLIP-2 stands out as having the highest cultural diversity, leveraging frozen pre-trained encoders (ViT-G \cite{fang2023eva} as the vision encoder and instruction-tuned FlanT5 \cite{chung2024scaling} as the language model) and a QFormer architecture. 

A particularly surprising finding is the robust performance of TCL with respect to both relevance and diversity -- despite being trained on a the smallest dataset among all models (4M images). TCL incorporates a unique uni-modal objective to make the model invariant to data modifications, 
which likely benefits the cross-modal alignment and joint multi-modal embedding learning. This may suggest that well-designed training objectives can sometimes compensate for smaller datasets, highlighting the significance of pre-training objectives alongside data scale. 

\paragraph{RQ$_2$: Do VLMs exhibit biases towards images from specific cultures?} From the full results in Appendix~\ref{appendix:results_countries} and \ref{appendix:results_regions} we can observe that there are no countries or regions that are consistently retrieved by models. A closer look reveals that the bias towards specific countries or regions is universal-specific. To demonstrate this point, we plot the top 5 retrieved images for 4 universal concepts, ``breakfast'',  ``funeral'', ``farming'', and ``wedding'', in Figure~\ref{fig:retrieval-analysis}. 

Despite exhibiting high country-level diversity and moderate region-level diversity, Figure~\ref{fig:retrieval-analysis} shows that the retrieved images for breakfast predominantly contain Western breakfast items such as eggs, sausages and toast. Similarly, the images for ``funeral'' mostly feature black dresses, and are overwhelmingly from Europe. With respect to ``farming'', CLIP and BLIP-2 mostly retrieve images from Western countries depicting technologically advanced farming tools and large green fields, whereas CoCA retrieves images from Africa and the Middle East of people working in the fields.   
Finally, the images for ``wedding'' are diverse across models, although CLIP focuses on more Western images whereas BLIP-2 prefers the Middle East (yet still retrieving images of white dresses). 

Despite being trained on large datasets, models like CLIP still exhibit notable biases towards Western cultures. 
While CoCA generally exhibits better diversity compared to CLIP and BLIP-2, all models display certain biases and preferences for Western-style elements, such as black dresses at funerals, white dresses at weddings, and eggs for breakfast. 

\paragraph{RQ$_3$: What are the challenges faced by VLMs in achieving high cultural diversity?}

A low diversity score may be attributed to various factors. First, the scarcity of images from non-Western cultures means that pre-training datasets are predominantly Western-centred \cite{shankar2017no}. Second, many large-scale pre-training datasets are predominantly sourced from Western-centric platforms, leading to the overrepresentation of Western cultures. Finally, typical pre-training objectives are designed to maximize general image-text alignment and do not specifically target cultural diversity, leading models to associate, for example, breakfast with eggs and weddings with white dresses.

\section{Task 2: Cultural Visual Grounding}
\label{sec:task2}
Visual grounding is essential for human-AI interactions, enabling users to reference regions using spatial cues and models to respond with precise visual answers, such as bounding boxes. Existing grounding datasets such as RefCOCO and its variants \cite{kazemzadeh-etal-2014-referitgame,yu2016modeling}, Flickr Entities \cite{plummer2015flickr30k}, Visual Genome \cite{krishna2017visual}, and GRIT \cite{gupta2022grit}  tend to focus on generic concepts and their images lack cultural contexts. 

\begin{table*}[!ht]
\centering
\scriptsize
\begin{tabular}{l|lccc|c}
\toprule
\textbf{Region} & \textbf{Country} & \textbf{Number of Concepts} & \textbf{Average bbox/image Ratio} & \textbf{Average Yolov5 Score} & \textbf{Human Eval (IoU)} \\ 
\midrule
Latin America & Argentina & 43 & 0.146 & 4.442 & 0.92\\  
 & Brazil & 32 & 0.153 & 3.906 & 0.87\\ 
 & Mexico & 43 & 0.163 & 5.744 & 0.91\\ \hline
North America & Canada & 26 & 0.118 & 5.500 & 0.92\\ \hline
East Asia & China & 39 & 0.163 & 4.106 & 0.94\\ 
 & South Korea & 41 & 0.151 & 5.317 & 0.87\\ \hline
South Asia & India & 53 & 0.112 & 5.698 & 0.88\\ 
 & Pakistan & 38 & 0.137 & 4.162 & 0.86\\ \hline
Middle-East Asia & Israel & 48 & 0.119 & 5.255 & 0.91\\ \hline
South East Asia & Philippines & 41 & 0.138 & 4.390 & 0.85\\  
 & Vietnam & 40 & 0.129 & 5.275 & 0.80\\ \hline
Africa & Nigeria & 36 & 0.137 & 3.611 & 0.92\\  
 & South Africa & 34 & 0.146 & 4.118 & 0.88\\ \hline
Europe & Poland & 40 & 0.216 & 3.150 & 0.95\\ 
 & Russia & 37 & 0.134 & 4.405 & 0.92\\ \hline
\end{tabular}
\caption{Detailed statistics of annotated images across different cultural groups and regions for Cultural Visual Grounding task.}
\label{tab:grounding-statistics}
\end{table*}

To address this limitation, we propose the task of  \textbf{Cultural Visual Grounding}, to evaluate the ability of VLMs to identify culture-specific concepts. We describe our dataset collection (Sec~\ref{sec:grounding:data}), the task and evaluation metric (Sec~\ref{sec:grounding:task}). We evaluate various models on our task (Sec~\ref{sec:grounding:baselines}), and report the performance in Sec~\ref{sec:grounding:results}. 


\subsection{Dataset Collection}
\label{sec:grounding:data}

\begin{table*}[t]
\centering
\scriptsize
\setlength{\tabcolsep}{2.5pt}
\begin{tabular}{llrll}
\toprule
\textbf{Model} & \textbf{Training Data} & \textbf{Data Size}  & \textbf{Vision Encoder}  & \textbf{LM} \\
\midrule
\textit{\textbf{Specialist Models}} \\
Grounding DINO \cite{liu2023grounding} & O365, GoldG, Cap4M & - & Swin-T (DINO) & BERT \\
\midrule
\textit{\textbf{Generalist Models}} \\
KOSMOS-2 \cite{peng2023kosmos} & LAION-2B, COYO, GRIT-91M & 2.8B & CLIP-ViT-L & Magneto \\
MiniGPT-v2 \cite{chen2023minigpt} & LAION, CC3M, SBU, GRIT-20M, VG, RefCOCO, VQA datasets & - & ViT & LLaMA-2-Chat-7B \\
QwenVL \cite{bai2023qwen} & LAION-en/zh, DataComp, COYO, CC, SBU, COCO & 1.4B & ViT-bigG & Qwen-7B \\
LLaVA-1.5 \cite{liu2024improved} & OKVQA, A-OKVQA, OCRVQA, TextCaps, VG, RefCOCO, GQA, ShareGPT & 1.2B & CLIP-ViT-L & Vicuna-13B  \\
\bottomrule
\end{tabular}
\caption{Overview of models benchmarked for the Cultural Visual Grounding task. \small **Note: Grounding DINO \cite{liu2023grounding} and MiniGPT-v2 \cite{chen2023minigpt} authors do not provide total training data size in the papers, so we leave that blank to avoid inaccurate numbers.}
\label{tab:models-grounding}
\end{table*}

\paragraph{Cultural Keywords.} In this task, we focus on 15 countries across 8 regions, detailed in Table~\ref{tab:grounding-statistics}. We extract from CANDLE 50 cultural keywords for each culture, covering topics such as food, rituals, clothing, etc. The list of keywords is detailed in Appendix~\ref{appendix:list_keywords_cvg}.

\paragraph{Images.} To obtain images corresponding to the keywords, we recruit annotators from the respective cultures through the CloudConnect Platform by Cloud Research.\footnote{\url{https://www.cloudresearch.com/}} We instructed annotators to find an image depicting the target cultural concept using Google Images. We emphasized that the images should be of high quality and do not solely depict the target concept but also include other visuals, to make sure the grounding task is not trivial. For instance, an image for the Korean sauce ``gochujang'' may contain gochujang along with other dishes. 

\paragraph{Bounding Boxes.} After selecting the images, annotators used a bounding box tool to draw a single bounding box (bbox) around the target concept. Each annotator was compensated \$50 USD for retrieving and annotating images for 50 concepts in their culture. 

\paragraph{Verification.} We perform an additional analysis step to verify that the cultural concept is not the main focus of the image. We do so by ensuring that the bbox-to-image ratio is less than 0.3. We also used an off-the-shelf object detection model, YOLOv5, to assess the number of objects in the image, filtering out images with fewer than 3 objects.\footnote{\scriptsize\url{https://pytorch.org/hub/ultralytics\_yolov5/}} Additionally, annotators were asked whether the concept was prevalent in their culture, and 1.3\% of the concepts were marked as not prevalent. This process resulted in the collection of 591 images. More detailed statistics of the collected data are provided in Table~\ref{tab:grounding-statistics}. 

Finally, we conduct a human evaluation to ensure quality by recruiting annotators from CloudConnect. Each annotator was asked to draw bounding boxes for the given cultural concept word. Annotator agreement was measured by calculating the Intersection over Union (IoU) score between the bounding boxes drawn by two different annotators. The IoU is calculated as: $IoU = \frac{|R_{\text{anno1}} \cap R_{\text{anno2}}|}{|R_{\text{anno1}} \cup R_{\text{\text{anno2}}}|}$.  Each annotator was compensated \$0.1 USD of each annotation. More detailed statistics of the collected data and human agreement scores (IoU) are provided in Table~\ref{tab:grounding-statistics}. 

\subsection{Task Definition and Evaluation Setup}
\label{sec:grounding:task}

Given an image $I$ and a query $q$ describing a cultural keyword, the goal is to predict a bounding box $R$ around the region in $I$ that corresponds to $q$. We evaluate models based on the overlap between the gold standard and predicted regions of interest, using Intersection over Union (IoU) as the metric: $IoU = \frac{|R \cap R_{\text{gold}}|}{|R \cup R_{\text{gold}}|}$. We consider a predicted bounding box correct if its IoU with the ground-truth bounding box is greater than 0.5, and report overall accuracy. It is crucial that models perform consistently well across different cultures.
 
\subsection{Models} 
\label{sec:grounding:baselines}

We benchmark a series of models on our grounding task, considering both \textit{specialist} models, designed explicitly for visual grounding tasks, and \textit{generalist} models, which can handle a wide range of vision-language tasks, such as captioning, question answering, and grounding. These models are listed in Table~\ref{tab:models-grounding}, along with their training data, vision and language backbones, and training methodology. 

The specialist model we include is Grounding DINO \cite{liu2023grounding}, a zero-shot object detection model that combines a Transformer-based detector \cite[DINO;][]{zhang2022dino} with phrase grounding pre-training \cite[GLIP;][]{li2022grounded}. The generalist models are multimodal large language models (MLLMs). MLLMs encode visual patches as tokens that a language model can understand. They perform visual grounding by generating bounding boxes in textual format, typically in the format of $
\langle X_{\text{left}} \rangle \langle Y_{\text{top}} \rangle \langle X_{\text{right}} \rangle \langle Y_{\text{bottom}} \rangle
$, denoting the coordinates of the top-left and bottom-right corners of the generated bounding box.

\begin{figure}[t]
     \centering\includegraphics[width=0.53\textwidth]{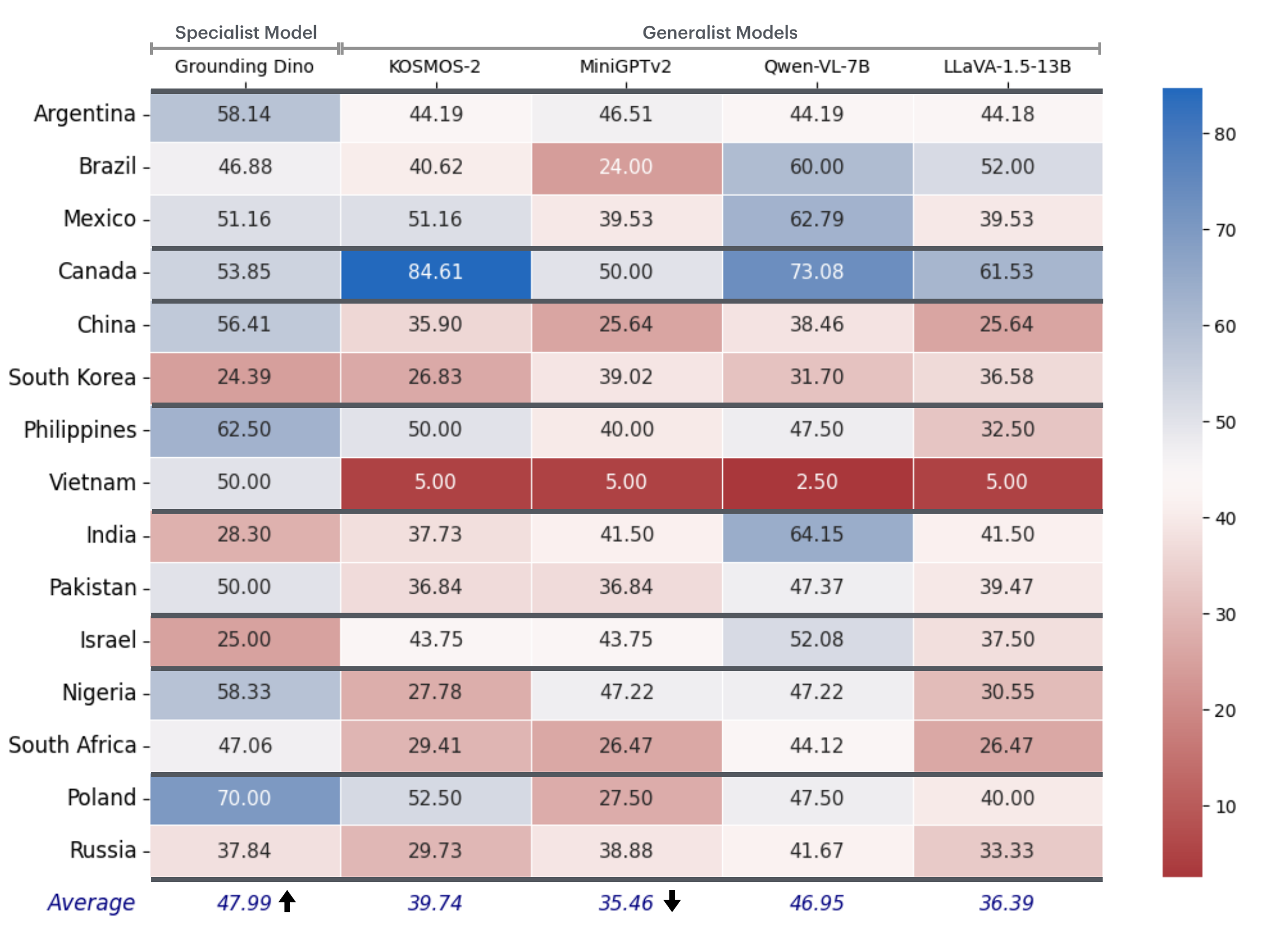}
     \caption{Country-level Accuracy of each model on the Cultural Visual Grounding task.} 
     \label{fig:culture-grounding-results}
\end{figure}

\begin{figure}[t]
     \centering
     \includegraphics[width=0.48\textwidth]{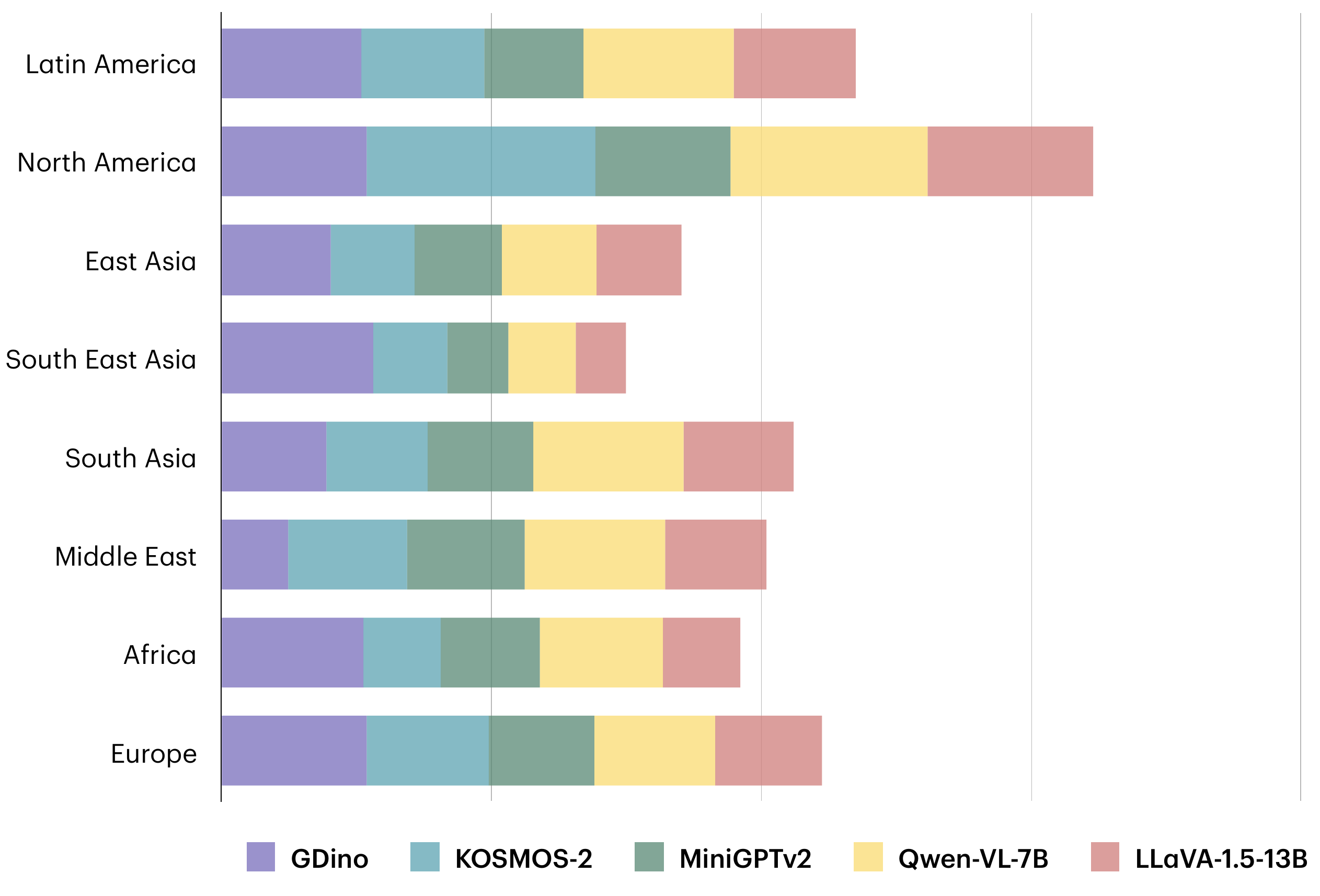}
     \caption{Culture group-level Accuracy for Cultural Visual Grounding.} 
     \label{fig:grounding-results}
\end{figure}

\subsection{Results and Analysis} 
\label{sec:grounding:results}

\begin{figure*}[t]
     \centering
     \includegraphics[width=\textwidth]{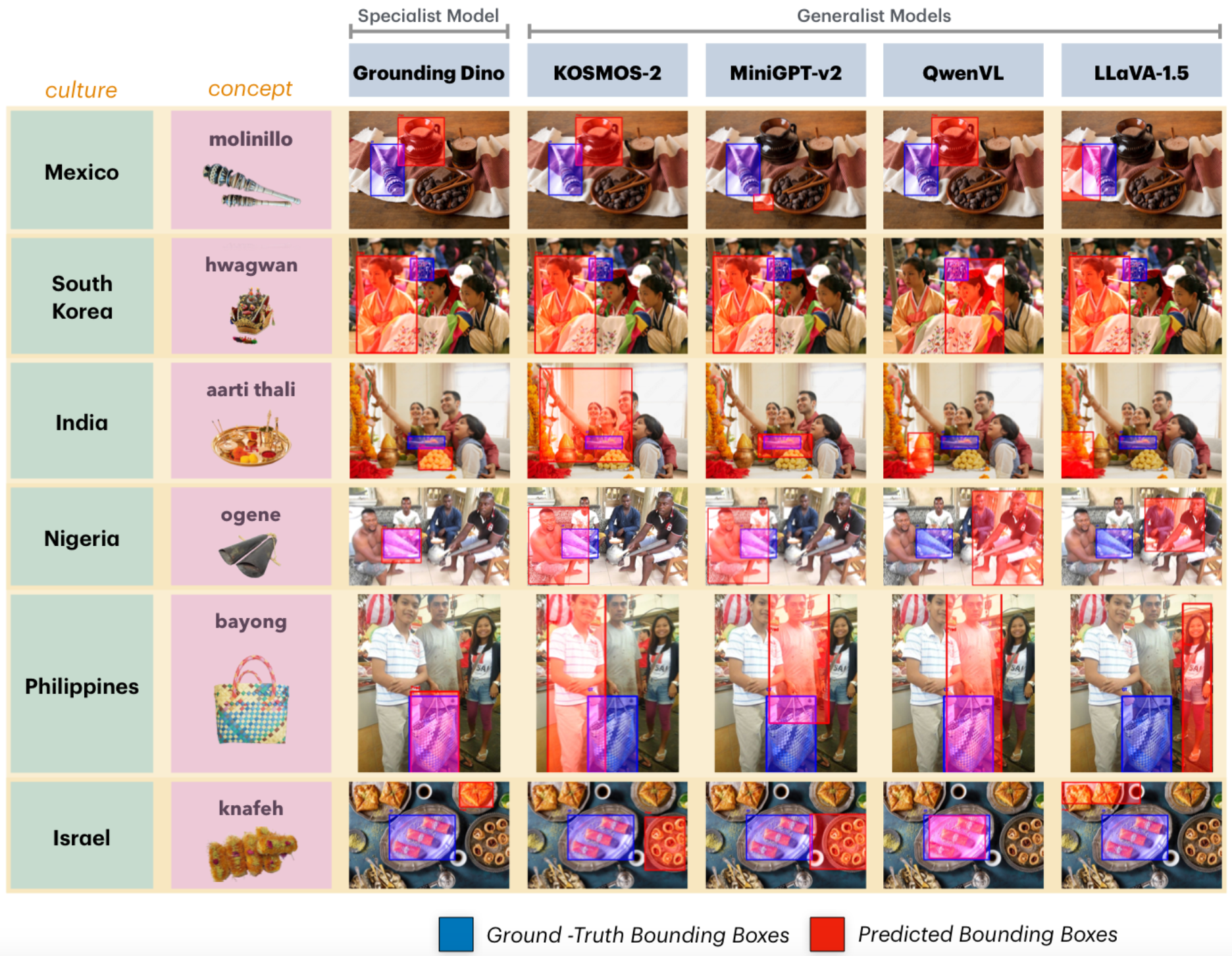}
     \caption{Qualitative Examples showing the performance of specialist and generalist models on Cultural Visual Grounding task.}
     \label{fig:grounding-failure-cases}
 \end{figure*}
 
\paragraph{RQ$_1$: Are VLMs able to identify culture-specific concepts?} 

Figure~\ref{fig:culture-grounding-results} presents the country-level accuracy of each model on the cultural visual grounding task. The overall performance across models is rather poor. Among all models, the specialist model Grounding DINO shows a relatively higher average performance (47.99\%) compared to the generalist models. 

Analyzing country-specific performance, we observe that KOSMOS-2 and QwenVL-7B exhibit strong accuracy in grounding elements for Canada and Mexico. Grounding DINO, on the other hand, performs well for Poland and the Philippines. All generalist models perform poorly on images from Vietnam, highlighting limited representation in training datasets. 

\paragraph{RQ$_2$: Do VLMs exhibit biases towards images from certain cultures?} To investigate whether VLMs show biases towards specific cultures, we plot the region-level performance for each model in Figure~\ref{fig:grounding-results}. We observe that almost all models achieve the highest performance on images from North America, with an average accuracy of 64.61\%, followed by a considerable drop in performance for images from Latin America (46.99\%) and Europe (44.49\%). This significant performance disparity may suggest that the VLMs were predominantly trained on images from North America.

Different models vary in their performances in the other regions. The generalist models show the most difficulty with images from South East Asia (accuracy between 18.75-27.5\%) and East Asia (31.11-35.08\%) while Grounding DINO performs worst on Middle Eastern images (25\%). 

\paragraph{RQ$_3$: What challenges do VLMs face in grounding culture-specific concepts?}

Figure~\ref{fig:grounding-failure-cases} presents some failure cases of the VLMs in the grounding task. We can categorize the errors into two primary types. In the first type, models draw a bounding box around an unrelated object. For example, in the image depicting a ``bayong'', a type of bag from the Philippines, the models frequently misidentify people as the ``bayong''. This suggests the model is unfamiliar with the term ``bayong'' and its visual representation. The other error type occurs when models draw the bounding box around another object with a shape similar to the target object. For instance, for ``ogene'', a double-bell instrument from Nigeria, some models incorrectly identified a person's arm as the ``ogene'', which may be due to shape similarity. This may suggest limited familiarity with the concept and its visual form.

\section{Conclusion}
\label{sec:conclusions}
In this work, we introduced a challenging benchmark, \benchmark{}, designed to evaluate the multicultural understanding of VLMs. 
\benchmark{} encompasses two tasks: retrieval of culturally diverse images depicting universal concepts and visual grounding of culture-specific concepts. Our findings from extensive experiments across a wide array of VLMs reveal significant performance variations across cultures, highlighting the existence of biases in current VLMs. 
Moving forward, future research should focus on collecting large-scale culturally diverse training datasets and devising training objectives that enhance models' representations of images from diverse cultures, ultimately paving the way for developing more inclusive and fair downstream applications.

\section*{Limitations}
\label{sec:limitations}
While our benchmark, \benchmark{}, provides a comprehensive evaluation of the multicultural understanding of VLMs, it is essential to acknowledge certain limitations as follows, 

\paragraph{Cultural Coverage.} Although our retrieval task encompasses 50 diverse cultures, the grounding task is restricted to only 15 cultures. This constraint arises from the availability of annotators on the crowdsourcing platform we used, Cloud Research. In future work, we aim to expand the grounding task to include a broader range of cultures.

\paragraph{Restricted cultural concepts.} Our study focuses on a selected set of cultural concepts or keywords from the CANDLE dataset. There might be more prominent cultural concepts that we could not cover. This limitation might restrict the comprehensiveness of our evaluation and overlook culturally significant aspects not captured by the selected keywords.

\paragraph{Metric for diversity.} We currently employ a diversity metric based on entropy to evaluate the cultural diversity of retrieved images. While this metric provides insights into the distribution of images across different cultures, it may not fully capture the nuanced variations in cultural representation. Our approach to regional diversity assessment may lack granularity, potentially overlooking finer distinctions in cultural diversity within regions. 

\section*{Ethical Consideration}
\label{sec:ethical}
\paragraph{Mapping from countries to regions.} For the purpose of our tasks, we mapped countries to broad regional categories as specified in Table 1. We acknowledge that cultures do not follow geographic boundaries and that this variation occurs at an individual level, shaped by one’s own life experiences. Despite this, we used our mapping as a practical starting point. This approach is a preliminary step, with the ultimate goal of developing systems that can learn from individual user interactions and adapt to diverse and evolving cultures.

\paragraph{Annotator selection and compensation} Annotators hired from Cloud Research were predominately based in USA, Canada, Australia, New Zealand, United Kingdom and Ireland. Participation was strictly limited to those who met specific criteria to maintain the relevance of the annotation process. Annotators were required to belong to a chosen ethnicity and to have lived in the designated countries for at least 5 of the past 15 years. This criterion ensured that participants had sufficient cultural context and lived experience relevant to the annotation tasks. We employed a second round of annotators for the human evaluation phase, ensuring none were repeated from the first round. 

\paragraph{Inadvertent stereotypes in collect images.} We recognize that some images used to capture cultural concepts might inadvertently perpetuate stereotypes. While our goal was to gather authentic cultural representations, we are aware of the ethical implications of including such content. We approached this task with the intention of collecting meaningful cultural data while being mindful of the potential for reinforcing harmful stereotypes.

\section{Acknowledgements}
\label{ack}
This work was funded, in part, by the Vector Institute for AI, Canada CIFAR AI Chairs program, Accelerate Foundation Models Research Program Award from Microsoft, an NSERC discovery grant, and a research gift from AI2.

\bibliography{anthology-part1,anthology-part2,custom}
\bibliographystyle{acl_natbib}

\appendix

\section{Appendix}
\label{sec:appendix}
\subsection{Complete Set of Results for Retrieval across Universals task}

\subsubsection{Results Across All Metrics}
\label{appendix:results_metrics}

Table \ref{tab:results-metrics-1} and \ref{tab:results-metrics-2} details results across all models. We show results for each universal and each metric. 

\begin{table*}[!ht]
\scriptsize
\resizebox{\textwidth}{!}{
\begin{tabular}{l|l|ccccccccccc}
\toprule
\textbf{Metric} & \textbf{Model} & \textbf{breakfast} & \textbf{clothing} & \textbf{dance} & \textbf{dessert} & \textbf{dinner} & \textbf{drinks} & \textbf{eating habits} & \textbf{farming} & \textbf{festival} & \textbf{funeral} \\
\midrule
\multirow{7}{*}{\textbf{Regional Diversity @ 10}} & CLIP & 65.35 & 69.9 & 65.35 & 65.35 & 63.88 & 69.9 & 73.65 & 69.9 & 47.29 & 65.05 \\
 & OpenCLIP & 73.65 & 69.9 & 73.65 & 73.65 & 79.67 & 79.67 & 63.88 & 67.62 & 40.97 & 63.88 \\
 & CoCA & 65.35 & 81.94 & 63.88 & 79.67 & 50.74 & 59.03 & 65.05 & 45.81 & 59.33 & 53.31  \\
 & TCL & 63.88 & 55.58 & 63.88 & 73.65 & 79.67 & 73.65 & 73.65 & 61.6 & 57.86 & 69.9  \\
 & ALBEF & 71.37 & 57.06 & 71.37 & 75.92 & 65.35 & 79.67 & 59.03 & 40.84 & 63.88 & 69.9  \\
 & BLIP-2 & 55.58 & 71.37 & 65.05 & 61.6 & 71.37 & 81.94 & 73.65 & 34.82 & 73.65 & 65.05  \\
 & FLAVA & 69.9 & 27.75 & 81.94 & 67.62 & 59.33 & 65.35 & 73.65 & 67.62 & 69.9 & 59.03 \\
\bottomrule
\end{tabular}}
\caption{First half of the results across all metrics and models for \textit{Retrieval Across Universals} task.}
\label{tab:results-metrics-1}
\end{table*}
\begin{table*}[!ht]
\scriptsize
\resizebox{\textwidth}{!}{\begin{tabular}{l|l|cccccccccc}
\toprule
\textbf{Metric} & \textbf{Model} & \textbf{greeting} & \textbf{headcoverings} & \textbf{instrument} & \textbf{lunch} & \textbf{marriage} & \textbf{music} & \textbf{religion} & \textbf{ritual} & \textbf{sports} & \textbf{transport} \\
\midrule
\multirow{7}{*}{\textbf{Regional Diversity@10}} & CLIP & 53.01 & 55.58 & 69.9 & 61.6 & 47.29 & 53.01 & 73.65 & 73.65 & 75.92 & 73.65 \\
 & OpenCLIP & 63.88 & 65.35 & 63.88 & 61.6 & 81.94 & 63.88 & 65.35 & 65.35 & 73.65 & 47.29 \\
 & CoCA & 73.65 & 71.37 & 53.31 & 55.58 & 63.88 & 59.03 & 75.92 & 75.92 & 71.37 & 73.65 \\
 & TCL & 79.67 & 79.67 & 69.9 & 63.88 & 50.74 & 73.65 & 34.82 & 63.88 & 65.35 & 75.92 \\
 & ALBEF & 69.9 & 73.65 & 73.65 & 67.62 & 73.65 & 40.84 & 73.65 & 53.01 & 67.62 & 44.72 \\
 & BLIP-2 & 73.65 & 57.06 & 75.92 & 73.65 & 50.74 & 69.9 & 50.74 & 67.62 & 39 & 53.01 \\
 & FLAVA & 65.35 & 75.92 & 63.88 & 81.94 & 44.72 & 67.62 & 73.65 & 81.94 & 69.9 & 69.9 \\
\bottomrule
\end{tabular}}
\caption{Second half of the results across all metrics and models for \textit{Retrieval Across Universals} task.}
\label{tab:results-metrics-2}
\end{table*}

\subsubsection{Results Across All Countries}
\label{appendix:results_countries}

Table \ref{tab:results-regions-1} and \ref{tab:results-regions-2} details the first 10 retrieved countries for each model and each universal.  

\begin{table*}[!ht]
\scriptsize
\resizebox{\textwidth}{!}{
\begin{tabular}{l|l|llllllllll}
\toprule
                           \textbf{Model} &  \textbf{First@10 Country}         & \textbf{breakfast} & \textbf{clothing} & \textbf{dance} & \textbf{dessert} & \textbf{dinner} & \textbf{drinks} & \textbf{eating\_habits} & \textbf{farming} & \textbf{festival} & \textbf{funeral} \\
\midrule
\multirow{10}{*}{CLIP}     & country 0 & germany            & italy             & italy          & fiji             & vietnam         & australia       & ethiopia                & hungary          & hungary           & china            \\
                           & country 1 & south africa       & sri lanka         & us             & thailand         & russia          & thailand        & netherlands             & italy            & hungary           & germany          \\
                           & country 2 & canada             & france            & france         & uk               & tunisia         & iran            & srilanka                & fiji             & sweden            & italy            \\
                           & country 3 & kenya              & greece            & australia      & phillippines     & ethiopia        & italy           & germany                 & poland           & singapore         & france           \\
                           & country 4 & australia          & fiji              & chile          & south africa     & portugal        & italy           & poland                  & japan            & new zealand       & uk               \\
                           & country 5 & somalia            & morocco           & canada         & new zealand      & us              & jamaica         & canada                  & ethiopia         & japan             & australia        \\
                           & country 6 & italy              & hungary           & uk             & hungary          & germany         & peru            & south korea             & lebanon          & greece            & us               \\
\bottomrule
\end{tabular}}
\caption{First half of the results for first 10 retrieved countries for \textit{Retrieval Across Universals} task.}
\label{tab:results-regions-1}
\end{table*}
\begin{table*}[!ht]
\scriptsize
\resizebox{\textwidth}{!}{
\begin{tabular}{llllllllllll}
\toprule
                           \textbf{Model} &  \textbf{First@10 Country} & \textbf{greetings} & \textbf{headcoverings} & \textbf{instrument} & \textbf{lunch} & \textbf{marriage} & \textbf{music} & \textbf{religion} & \textbf{ritual} & \textbf{sports} & \textbf{transport} \\
\midrule
\multirow{10}{*}{CLIP}     & country 0 & vietnam           & tunisia                & hungary             & vietnam        & portugal          & somalia        & thailand          & thailand        & thailand        & ethiopia           \\
                           & country 1 & thailand          & morocco                & netherlands         & sweden         & germany           & russia         & china             & srilanka        & italy           & peru               \\
                           & country 2 & us                & greece                 & sweden              & indonesia      & brazil            & turkey         & peru              & germany         & us              & china              \\
                           & country 3 & south korea       & china                  & vietnam             & portugal       & hungary           & peru           & jamaica           & thailand        & hungary         & peru               \\
                           & country 4 & thailand          & morocco                & brazil              & peru           & greece            & netherlands    & china             & china           & fiji            & kenya              \\
                           & country 5 & vietnam           & egypt                  & chile               & phillippines   & australia         & hungary        & tanzania          & ethiopia        & japan           & thailand           \\
                           & country 6 & indonesia         & hungary                & iran                & germany        & canada            & uk             & kenya             & morocco         & turkey          & india              \\
\end{tabular}} 
\caption{Second half of the results for first 10 retrieved countries for \textit{Retrieval Across Universals} task.}
\label{tab:results-regions-2}
\end{table*}

\subsubsection{Results Across All Regions}
\label{appendix:results_regions}

Table \ref{tab:results-cultures-1} and \ref{tab:results-cultures-2} details the first 10 retrieved regions for each model and each universal.  

\begin{table*}[!ht]
\scriptsize
\resizebox{\textwidth}{!}{
\begin{tabular}{llllllllllll}
\toprule
                           \textbf{Model} &  \textbf{First@10 Regions} & \textbf{breakfast} & \textbf{clothing} & \textbf{dance}   & \textbf{dessert} & \textbf{dinner}  & \textbf{drinks}  & \textbf{eating\_habits} & \textbf{farming} & \textbf{festival} & \textbf{funeral} \\
\midrule
\multirow{10}{*}{CLIP}     & region 0 & Europe             & Europe            & Europe           & Oceania          & Southeast Asia   & Oceania          & Africa                  & Europe           & Europe            & East Asia        \\
                           & region 1 & Africa             & South Asia        & North America    & Southeast Asia   & Europe           & Southeast Asia   & Europe                  & Europe           & Europe            & Europe           \\
                           & region 2 & North America      & Europe            & Europe           & Europe           & Africa           & Middle East Asia & South Asia              & Oceania          & Europe            & Europe           \\
                           & region 3 & Africa             & Europe            & Oceania          & Southeast Asia   & Africa           & Europe           & Europe                  & Europe           & Southeast Asia    & Europe           \\
                           & region 4 & Oceania            & Oceania           & Latin America    & Africa           & Europe           & Europe           & Europe                  & East Asia        & Oceania           & Europe           \\
                           & region 5 & Africa             & Africa            & North America    & Oceania          & North America    & Caribbean        & North America           & Africa           & East Asia         & Oceania          \\
\bottomrule
\end{tabular}}
\caption{First half of the results for first 10 retrieved regions \textit{Retrieval Across Universals} task.}
\label{tab:results-cultures-1}
\end{table*}
\begin{table*}[!ht]
\scriptsize
\resizebox{\textwidth}{!}{
\begin{tabular}{llllllllllll}
\toprule
                           \textbf{Model} &  \textbf{First@10 Regions} & \textbf{greeting} & \textbf{headcoverings} & \textbf{instrument} & \textbf{lunch}   & \textbf{marriage} & \textbf{music}   & \textbf{religion} & \textbf{ritual}  & \textbf{sports}  & \textbf{transport} \\
\midrule
\multirow{10}{*}{CLIP}     & region 0 & Southeast Asia    & Africa                 & Europe              & Southeast Asia   & Europe            & Africa           & Southeast Asia    & Southeast Asia   & Southeast Asia   & Africa             \\
                           & region 1 & Southeast Asia    & Africa                 & Europe              & Europe           & Europe            & Europe           & East Asia         & South Asia       & Europe           & Latin America      \\
                           & region 2 & North America     & Europe                 & Europe              & Southeast Asia   & Latin America     & Middle East Asia & Latin America     & Europe           & North America    & East Asia          \\
                           & region 3 & East Asia         & East Asia              & Southeast Asia      & Europe           & Europe            & Latin America    & Caribbean         & Southeast Asia   & Europe           & Latin America      \\
                           & region 4 & Southeast Asia    & Africa                 & Latin America       & Latin America    & Europe            & Europe           & East Asia         & East Asia        & Oceania          & Africa             \\
                           & region 5 & Southeast Asia    & Middle East Asia       & Latin America       & Southeast Asia   & Oceania           & Europe           & Africa            & Africa           & East Asia        & Southeast Asia     \\
\bottomrule
\end{tabular}}
\caption{Second half of the results for first 10 retrieved regions \textit{Retrieval Across Universals} task.}
\label{tab:results-cultures-2}
\end{table*}

\subsection{List of Cultural Keywords in Cultural Visual Grounding dataset}
\label{appendix:list_keywords_cvg}

Table \ref{tab:retrieval-cultures-list} lists the cultural concepts for each country in the Cultural Visual Grounding Dataset.

\begin{table*}[!ht]
\centering
\scriptsize
\resizebox{\textwidth}{!}{
\begin{tabular}{lp{15cm}}
\toprule
\textbf{Country} & \textbf{Cultural Concepts} \\
\midrule
\textbf{Argentina}               & alfajor, alpargatas, asado, bandoneon, bifes a la criolla, boina, bolero, bombilla, carbonada, chimichurri, chipa, chocotorta, choripan, churros rellenos, dulce de batata, dulce de leche, dulce de membrillo, empanada, facturas, faina, gaucho knife, humita, locro, lomito sandwich, malbec, matambre, mate, medialuna, milanesa, morcilla, parrilla, pascualina, pastel de papa, pebete, picada, provoleta, rabanito, ravioles, rosca de pascua, sandwich de miga, torta frita, vino patero, yerba  \\
\textbf{Brazil}         & acai, acaraje, alfajor, baiao, bombacha, bumba-meu-boi, brigadeiro, cachaca, caipirinha, carimbo, chimarrao, churrasco, cocar, cuica, empada, espetinho, farofa, feijoada, frescobol, moqueca, pacoca, pao de queijo, rapadura, requeijao, rosca, romeu e julieta, samba, sarongue, tapioca, tucupi, vatapa \\
\textbf{Canada}              & bagel, bannock, beavertail pastry, blueberry grunt, butter tart, caribou, cipaille, caesar cocktail, cretons, date squares, donair, flipper pie, garlic fingers, inukshuk, jiggs dinner, maple taffy, nanaimo bar, peameal bacon, pemmican, persian roll, poutine, rappie pie, sugar pie, toboggan, toque, tourtiere \\
\textbf{China}        & baozi, bianlian, bianzhong, biang biang noodles, chinese knot, chinese lantern, chinese seal, cong you bing, doufu, dragon beard candy, erhu, fenghuang crown, gongbi, guzheng, hongbao, hotpot, huanghuali furniture, hulusi, jiaozi, jinghu, laziji, liuli, longjing tea, luo han guo, mala tang, mahjong tiles, mooncake, paper cutting, peking opera mask, pipa, qipao, shengjianbao, suzhou embroidery, wushu sword, xiao long bao, xun, yuanyang hotpot, zongzi \\
\bottomrule
\end{tabular}}
\caption{List of cultures concepts covered in Cultural Visual Grounding dataset}
\label{tab:retrieval-cultures-list}
\end{table*}

\subsection{Model Checkpoints}

\begin{itemize}
    \item \textbf{CLIP}: laion/CLIP-ViT-g-14-laion2B-s12B-b42K
    \item \textbf{OpenCLIP}: clip-vit-base-patch32
    \item \textbf{Coca} CoCa-ViT-B-32-laion2B-s13B-b90k
    \item \textbf{llava}: llava-hf/llava-1.5-13b-hf
    \item \textbf{Qwen}: Qwen/Qwen-VL-Chat
    
\end{itemize}


\end{document}